\documentclass[letterpaper, 10 pt, conference]{ieeeconf}  % Comment this line out if you need a4paper

\IEEEoverridecommandlockouts                              % This command is only needed if 
\usepackage{graphics} % for pdf, bitmapped graphics files
\usepackage{times}
\usepackage{soul}
\usepackage{url}
\usepackage[hidelinks]{hyperref}
\usepackage[utf8]{inputenc}
\usepackage[small]{caption}
\usepackage{graphicx}
\usepackage{amsmath}
\usepackage{amssymb}
\usepackage{booktabs}
\usepackage[noend]{algpseudocode}
\usepackage{multirow}
\usepackage{algorithm}
\usepackage{xcolor}

\title{Policy Optimization to Learn Adaptive Motion Primitives in Path Planning with Dynamic Obstacles}
\author{Brian Angulo$^{1}$, Aleksandr Panov$^{2}$ and Konstantin Yakovlev$^{3}$% <-this % stops a space
\thanks{$^{1}$Brian Angulo is with Moscow Institute of Physics and Technology and JSC Integrant
        {\tt\small brian.angulo@phystech.edu}}
\thanks{$^{2}$Aleksandr I. Panov is with Federal Research Center for Computer Science and Control RAS and AIRI
        {\tt\small panov@airi.net}}%
\thanks{$^{3}$Konstantin Yakovlev is with Federal Research Center for Computer Science and Control RAS, Moscow Institute of Physics and Technology, AIRI
        {\tt\small yakovlev.ks@gmail.com}}
\thanks{Accepted as regular paper to Robotics and Automation Letters in December 2022.}
}

\begin{document}

\maketitle

\begin{abstract}

This paper addresses the kinodynamic motion planning for non-holonomic robots in dynamic environments with both static and dynamic obstacles -- a challenging problem that lacks a universal solution yet. One of the promising approaches to solve it is decomposing the problem into the smaller sub-problems and combining the local solutions into the global one. The crux of any planning method for non-holonomic robots is the generation of motion primitives that generates solutions to local planning sub-problems. In this work we introduce a novel learnable steering function (policy), which takes into account kinodynamic constraints of the robot and both static and dynamic obstacles. This policy is efficiently trained via the policy optimization. Empirically, we show that our steering function generalizes well to unseen problems. We then plug in the trained policy into the sampling-based and lattice-based planners, and evaluate the resultant POLAMP algorithm (Policy Optimization that Learns Adaptive Motion Primitives) in a range of challenging setups that involve a car-like robot operating in the obstacle-rich parking-lot environments. We show that POLAMP is able to plan collision-free kinodynamic trajectories with success rates higher than 92\%, when 50 simultaneously moving obstacles populate the environment showing better performance than the state-of-the-art competitors. 

The code is available at \url{https://github.com/BrianAnguloYauri/POLAMP}.

\end{abstract}

\section{Introduction}

Autonomous robotic systems have become one of the most popular research topics in recent years due to its huge potential social benefits. In particular, autonomous driving is developing rapidly and at the same time requires efficient motion planning in complex and highly dynamic environments, meanwhile taking into account the kinodynamic constraints of an non-holonomic autonomous vehicle. Often, the planners that address the first aspect of the problem, i.e. dynamic environment, like the ones presented in~\cite{Otte2016RRTX, phillips2011sipp} do not take into account the kinodynamic constraints. On the other hand, kinodynamic planners often do not explicitly reason about the future changes of the environments, even if these changes are for-seen, e.g. predicted by the control system of the robot. In this work we want to enrich the kinodynamic planning methods with the ability to take the dynamics of the environment as well (at the planning stage).

Two common approaches to kinodynamic planning are widespread: lattice-based and sampling-based planning methods. Lattice-based planning methods utilize the so-called motion primitives~\cite{MotionPrimitives} that form a regular lattice. Each motion primitive represents a small segment of kinodynamically feasible trajectory of the robot, which is pre-computed before planning. At the planning stage the search-based algorithms (e.g. A*~\cite{hart1968formal} of its variants) are used to find the resultant trajectory, represented as a sequence of the motion primitives. Contrary, sampling-based planners, e.g. RRT~\cite{RRT} or RRT*~\cite{RRT*}, grow a search tree by sampling states in the robot's configuration space and invoke a local planner to connect two states while respecting the kinematic constraints of the robot. Thus, the motion primitives are constructed online (i.e. while planning).
%Thus these planners create the motion primitives online while planning.
%However both of them do not consider the sensor information of the robot while planning.
\begin{figure}[]
    \centering
    \includegraphics[width=0.49\textwidth]{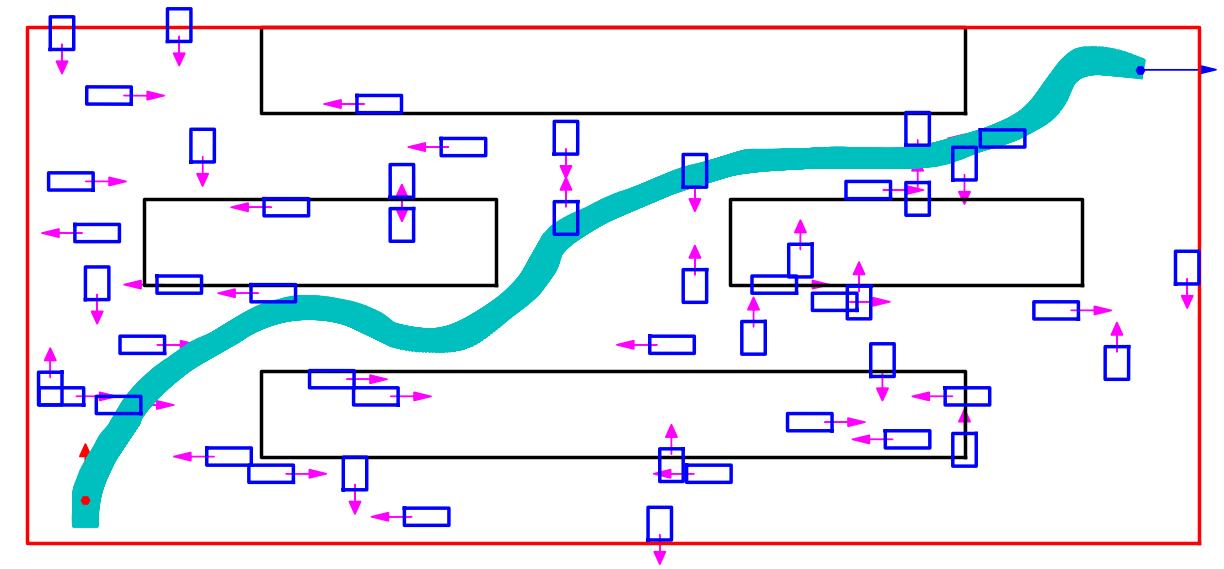}
    \caption{Illustration of the POLAMP algorithm. \textbf{Red} arrow represent the start state, \textbf{Blue} arrow -- the goal one. \textbf{Black} and \textbf{Blue} rectangles represent the static and dynamic obstacles respectively. \textbf{Cyan} curve is the generated trajectory by A*-POLAMP.}\label{fig:IntroFigure}
\end{figure}

One of the prominent approaches to alleviate the complexity of local planners to respect the kinematic constraints of the robot is to use methods based on reinforcement learning such as methods proposed in RL-RRT~\cite{Chiang2019RL-RRT} and PRM-RL~\cite{Faust2018PRM-RL}. In this work, we suggest Policy Optimization algorithm to Learn Adaptive Motion Primitives (POLAMP) to take into account the future changes of the environment at the planning stage, while producing plans that satisfy the kinodynamic constraints of the robot. POLAMP utilizes a reinforcement learning approach to find a policy that generates local segments of trajectory which are embedded in the global planning algorithms, RRT and A*, to generate a global motion plan. Our learnable local planner utilizes local observation to avoid both static and dynamic obstacles and, as well, respect the kinodynamic constraints of the robot. As a result, POLAMP is able to generate feasible solutions with high success rate ($>92\%$) in the environments with up to $50$ moving obstacles thus outperforming competitors.

%\end{enumerate}

\section{Related Work}

The problem of kinodynamic planning is well researched and various approaches such as graph-based, sampling based, optimization, reinforcement learning or the combination of them are used, see~\cite{gonzalez2015review} for review. Nevertheless, the kinodynamic planning in presence of dynamic obstacles is still a challenging problem.

A widespread approach to kinodynamic planning in robotics is sampling-based planners. The most popular way to account for the robot's dynamics is to sample in the robot's state space and attempt to connect states via different local planners~\cite{hwan2011anytime,webb2013kinodynamic,BIT*SQP,Otte2016RRTX}, including for car-like robot~\cite{vailland2021CubicBezier}. The method described in this work also relies on the local planner, but it is learnable and takes the moving obstacles into account. Unlike the methods presented in~\cite{Otte2016RRTX, Chen2019Horizon}, it assumes that the information on how the obstacles are intended to move in future is available (e.g. predicted from the sensors' observations) and takes this information into account while planning.

Recently a lattice-based planner for car-like robots in highly dynamic environments was proposed~\cite{lin2021search}. Other variants of lattice-based planners for car-like robots are described in ~\cite{MotionPrimitives, rufli2010design, ziegler2009spatiotemporal}. Contrary to these algorithms the suggested method does not construct a lattice in the high-dimensional space to search for a feasible plan, but uses a local learnable planner to connect states.

There also exist methods that, first, generate a rough path, often the one that does not take the kinodynamic constraints into account, and then generate controls to follow the path respecting the system's dynamics and avoiding obstacles. The variants of these methods are described in~\cite{perez2021robot, kontoudis2019kinodynamic}. Unlike them the method proposed in this work builds a feasible trajectory in one planning step. Avoiding the moving obstacles is performed by utilizing the knowledge of their future trajectories.
%There are other works which address the kinodynamic motion planning using Actor-Critic architecture with completely unknown dynamic system. Unlike these methods, instead of following through initial guess generating kinodynamic controls primarily we construct a kynodynamic path between the given states. We can add other method.}

Finally, the most similar methods to the one presented in this article are RL-RRT~\cite{Chiang2019RL-RRT} and PRM-RL~\cite{Faust2018PRM-RL}. Our method also uses a learning local planner inside a sampling-based planner. However, unlike these methods our local planner considers the presence of dynamic obstacles.

\section{Problem Statement}

We are interested in planning a feasible kinodynamic trajectory for a non-holomonic robot, that avoids both static and moving obstacles. In particular we are interested in car-like robots whose dynamics is described as~\cite{surveyMotionPlanning}:

\begin{align}\label{eq:diffEquationsRobot}
    &\dot{x} = v cos(\theta)\nonumber\\
    &\dot{y} = v sin(\theta)\\ 
    &\dot{\theta} = \frac{v}{L} \tan(\gamma), \nonumber
\end{align}
where $x$,$y$ are the coordinates of the robot's reference point (middle of the rear axle), $\theta$ is the orientation, $L$ is the wheel-base, $v$ is the linear velocity, $\gamma$ is the steering angle. The former three variables comprise the state vector: $\boldsymbol{x}(t)=(x,y,\theta)$.
The latter two variables form the control vector: $\boldsymbol{u}(t) = (v, \gamma)$, which can also be re-written using the acceleration $a$ and the rotation rate $\omega$ as follows: $v = v_0 + a \cdot t, \gamma = \gamma_0 + \omega \cdot t$.

 The robot is operating in the 2D workspace populated with static and dynamic obstacles. Their shapes are rectangular (as the one of the robot). Let $Obs = \{ Obs_1(t), ..., Obs_n(t)\}$ denote the set of obstacles, where $Obs_i(t)$ maps the time moments to the positions of the obstacle's reference point in the workspace. For the static obstacles it obviously holds that $\forall t:Obs_i(t)=Obs_i(0)$. In our work, we consider the functions $Obs_i(t)$ to be known. 
 
 Denote by $\mathcal{X}_{free}(t)$ all the configurations of the robot which are not in collision with any of the obstacles at time moment $t$ (w.r.t. the robot's and the obstacles' shapes). The problem now is to find the controls (as functions of time) that move the robot from its start configuration $s_{start}$ to the goal one $s_{goal}$ s.t. that the kinodynamic constraints~(\ref{eq:diffEquationsRobot}) are met and the resultant trajectory lies in $\mathcal{X}_{free}(t)$. %In this work we assume the time to be discretized into the time steps: $T={0,1,...,T_{max}}$, with $T_{max}$ being the time horizon, i.e. the robot has to reach the goal no later than $T_{max}$.

\section{Method}

We rely on the combination of the global and the local planners to solve the described problem. Global planner is aimed to systematically decompose the problem into the set of sub-problems which are easier to solve, i.e. moving from one configuration to another. The local planner is tailored to solve the latter problem. Any such a sub-problem is in essence the two-boundary value problem with additional constraints (prohibiting the robot to collide with both static and dynamic obstacles) which is hard to solve directly. The crux of our approach is to cast this problem as the partially-observable Markov decision process (POMDP) and to obtain the policy for solving the POMDP via the reinforcement learning, and more specifically via the custom-tailored Proximal Policy Optimization algorithm. Once the policy is obtained (learned) we plug it into the global planner. As a global planner we can use an adaptation of the renowned algorithms, RRT and A*, to get the final solver. We name this type of solvers as POLAMP -- Policy Optimization to Learn Adaptive Motion Primitives.

\begin{figure}[t]
    \centering
    \includegraphics[width=0.4\textwidth]{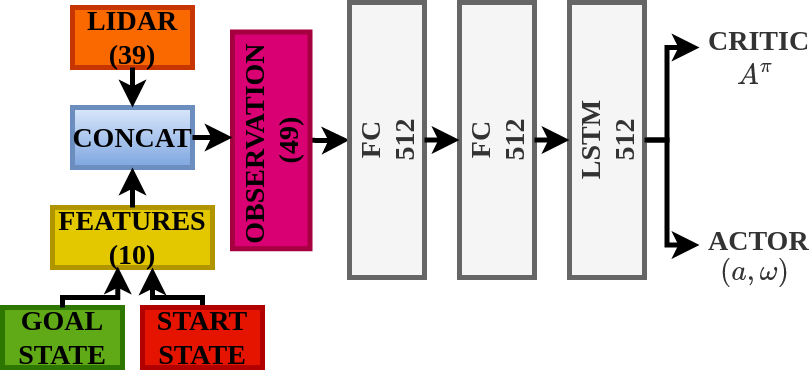}
    \caption{Actor-Critic architecture that is implemented in POLAMP}
    \label{fig:actor-critic-arch}
\end{figure}

\subsection{Learnable Local Planner}

\paragraph{Background}

Formally, POMDP can be represented as a tuple $(\mathcal{S}, \mathcal{A}, \mathcal{P}, \mathcal{R}, \Omega)$, where $\mathcal{S}$ is the state space, $\mathcal{A}$ is the action space, $\mathcal{P}$ is the state-transition model, $\mathcal{R}$ is the reward function, $\Omega$ is the observation space. During learning at each time step the agent receives an observation $o_t \in \Omega$, takes an action $a_t \in \mathcal{A}$ and receives a reward $r_t \in \mathcal{R}$. The goal is to learn a policy, i.e. the mapping from the observations to the distributions of actions, $\pi: \Omega\rightarrow P(\mathcal A)$. The policy should maximize the following expected return from the start state $s_t$: 
\begin{equation*}
   J(\pi) = \mathbb E_{r_{i}, s_{i} \sim \mathcal{P}, a_{i} \sim \pi}[\sum_{i=t}^T \gamma^{i-t}r(s_i, a_i) | s_t, a_t], i>t,
\end{equation*}
where $\gamma$ is the discounting factor.

The $Q$-function is used to concise definition of the most essential information for the agent in order to make an optimal decision:
\begin{equation*}
    Q^{\pi}(s_t, a_t) = \mathbb{E}_{r_{i}, s_{i} \sim \mathcal{P}, a_{} \sim \pi}[R_t | s_t, a_t], i>t,
\end{equation*}

\begin{figure}[t]
    \centering
    \includegraphics[width=0.35\textwidth]{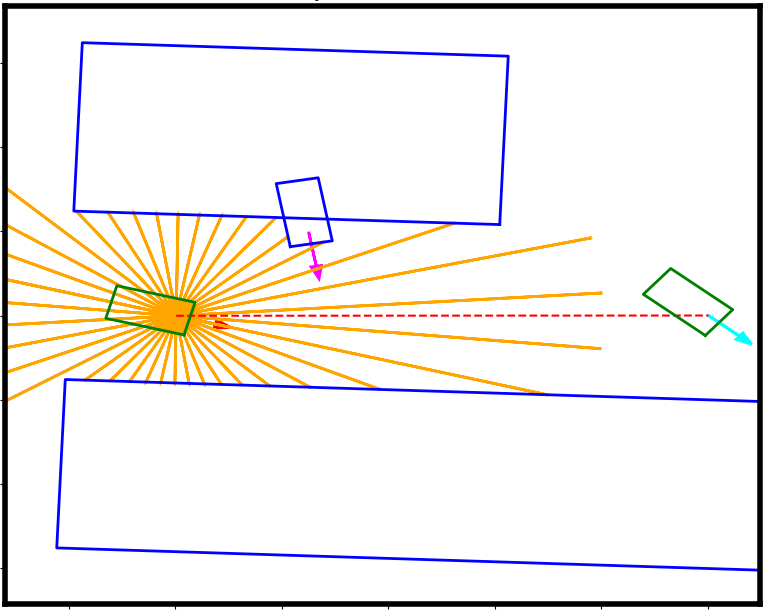}
    \caption{The learning environment. \textbf{Green} rectangle with the \textbf{Red} arrow is the current state of the robot. \textbf{Green} rectangle with the \textbf{cyan} orientation is the goal desired state. \textbf{Blue} rectangles are the static obstacles and the \textbf{Blue} rectangle with \textbf{Pink} arrow is the moving obstacle. \textbf{Orange} lines are the laser beams.}
    \label{fig:learningEnvironment}
\end{figure}

In this paper, we consider algorithms of the actor-critic family, which are more stable, have less variance, and are less prone to convergence to a local minimum. The actor updates the policy approximator $\hat \pi_w$ using the following equation:
\begin{equation*}
    \nabla_w J(\hat{\pi}_w) = \mathbb{E}_{\hat{\pi}_w} \big[ \nabla_w \log \hat{\pi}_w(s,a) Q^{\hat{\pi}_w}(s,a) \big],
\end{equation*}
where $\hat \pi_w$ is an arbitrary differentiable policy. Critic evaluates the approximation of the $Q^{\hat{\pi}_w}(s,a)$ value for the current policy $\hat \pi_w$. Actor-critic algorithms have two sets of parameters: a critic updates parameters $\phi$ of the $Q$-function, and an actor updates parameters $w$ of the policy according to the critic assumptions.

%Indeed it is possible to use various actor-critic learning algorithms, such as TD3~\cite{TD3}, SAC~\cite{SAC}, and PPO~\cite{PPO}, as the local planner in our learning environment. 

%owever, in our preliminary experiments  %
In this work, we use Proximal Policy Optimization method~\cite{PPO} (PPO) because it has shown the best performance among other methods in our preliminarly evaluation. Actor part of the PPO optimizes the clipped loss function
\begin{multline*}
    L(s, a, w_k, w) = \min (\frac{\pi_w(a|s)}{\pi_{w_k}(a|s)} A^{\pi_{w_{old}}}(s, a), \\
    clip(\frac{\pi_w(a|s)}{\pi_{w_{old}}(a|s)}, 1- \epsilon, 1+\epsilon) A^{\pi_{w_{old}}}(s, a)),
\end{multline*}
where $A^{\pi_{w}}$ is an estimation of the advantage function $A(s, a) = Q(s, a) - V(s)$ given by the critic part. Clipping is a regularizer removing incentives for the policy to change dramatically. The hyperparameter $\epsilon$ corresponds to how far away the new policy can go from the old while still profiting from the objective. When integrating the PPO algorithm into our method, we considered the state $s_t$ as a function from observation $s_t\approx f(o_t)$, where $f$ is lower layers of neural network approximator of the actor and critic shown in Fig.~\Ref{fig:actor-critic-arch}.

\paragraph{Observations, actions and rewards\label{localPlannner}}\label{paragraph:observations-actions-reward}

In this paper, we consider actions $a_t = (a, \omega) \in R^2$ that are composed of setting the linear acceleration $a \in (-5, 5) \; m/s^2$ and rotation rate $\omega \in (-\pi/12, \pi/12) \;rad/s$. The latter ones can be converted to robot's controls using the transformations for Eq.~\ref{eq:diffEquationsRobot}, where we set the range of the linear velocity in $v \in (0, 4) \; m/s$ and steering angle in $\gamma \in (-\pi/6, \pi/6) \;rad$.

The observation $o_t$ is a vector that consists of the $N_{beams}=39$ measurements of the lidar that cover the 360$^{\circ}$ surrounding of the robot up to the length of $beam_{max} = 20\;m$ -- see Fig.~\ref{fig:learningEnvironment} concatenated with the features  $(\Delta x, \Delta y, \Delta \theta, \Delta v, \Delta \gamma, \theta, v, \gamma, a, \omega)$, where $\Delta(s_i)$ stands for the difference between the respective parameter $s_i$ of the goal state and the current one, $(\theta, v, \gamma)$ are last three parameters of the current state and $(a, \omega)$ are the current controls. We consider an ideal environment, so both simulation and actuation model do not have errors.

\begin{figure*}[t]
    \centering
        \includegraphics[width=0.32\textwidth]{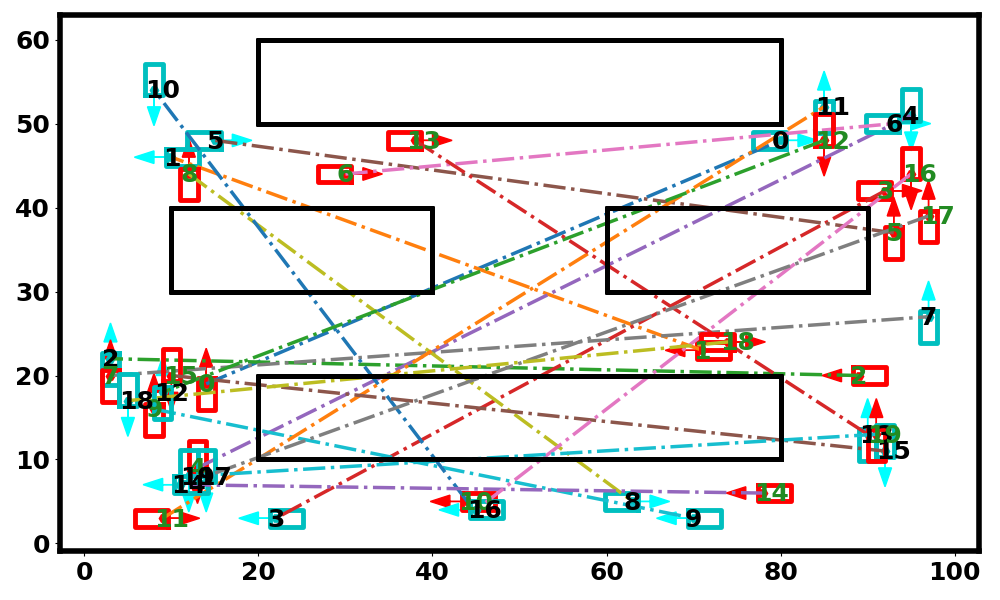}
        \includegraphics[width=0.32\textwidth]{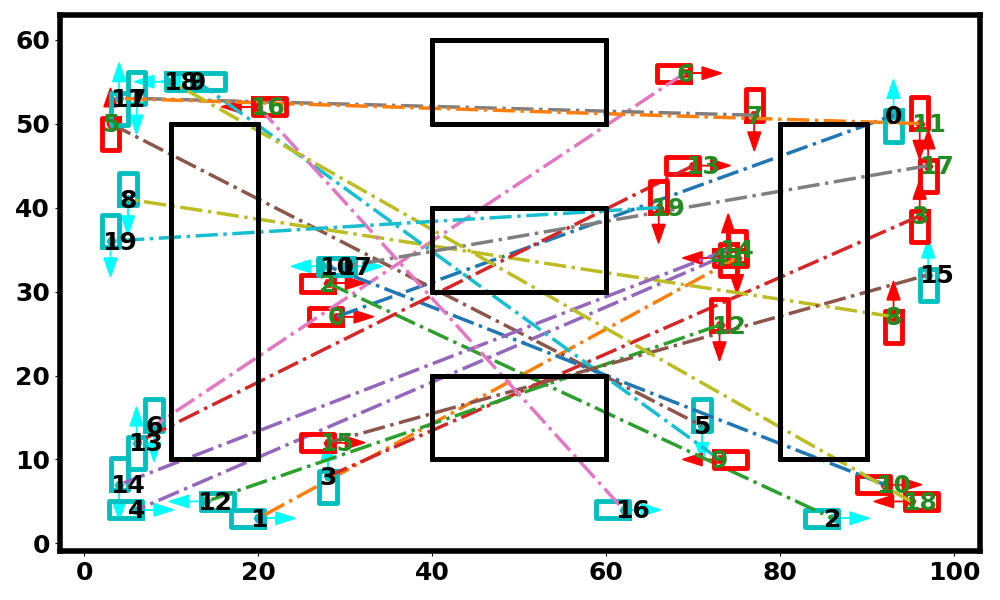}
        \includegraphics[width=0.32\textwidth]{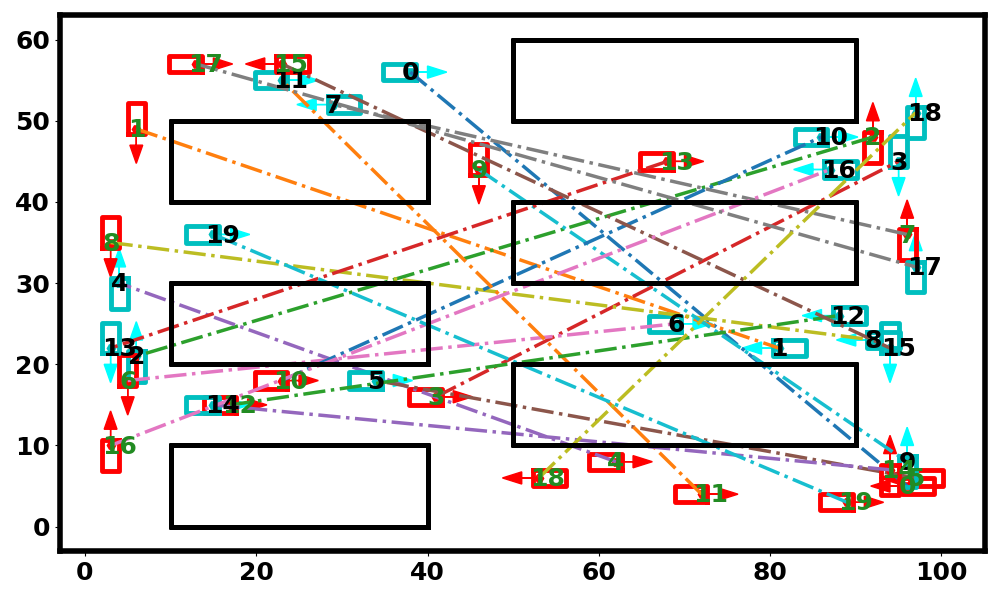}
    \caption{Maps (1-3) used in our tests. \textbf{Red} rectangles and arrows show the start coordinates and orientations (with the start velocity $v = 0$ and the steering angle $\gamma = 0$), and \textbf{Cyan} rectangles and arrows show the goal coordinates and orientations.}
    \label{fig:environments}
\end{figure*}

The reward function is described by:
\begin{equation*}
    \mathcal R = w_r^T [r_{\text{goal}}, r_{\text{col}}, r_{\text{field}}, r_{t}, r_{\text{backward}}, r_{v_\text{max}}, r_{\gamma_\text{max}}],
\end{equation*}

where $w_r$ is a vector of weights, $r_{\text{goal}}$ is 1 if the agent has reached the goal state with the $(\epsilon_{\rho}, \epsilon_{\theta})$ tolerance and 0 otherwise, $r_{\text{col}}$ is $-1$ if the agent collides with the obstacles and 0 otherwise, $ r_{\text{field}} = \rho_{curr} - \rho_{last}$, where $\rho_{last} = \|s_{t - 1} - s_{goal}\|$ and $\rho_{curr} = \|s_t - s_{goal}\|$ we penalize the agent for moving away from the goal, $r_{t}=-1$ is the constant penalty for each time step, $r_{\text{backward}}$ is $-1$ the the agent is using rear gear (moving backwards) and  0 otherwise, $r_{v_\text{max}}$ is $-1$ for exceeding the maximum speed limit, $r_{\gamma_\text{max}}$ is $-1$ for exceeding the maximum of steering angle threshold. We set the weights to be $w_r = [20, 8, 1, 0.1, 0.3, 0.5, 0.5]$ (empirically those values result in a more efficient learning).

\paragraph{Curriculum policy learning\label{policyLearning}}
To accelerate training end we propose a three-stage curriculum learning (see Fig.~\ref{fig:curriculum-training}). During the first stage, we train the agent in the empty environment. This stage is tailored to learn the kinodynamic constraints of the vehicle. Once the agent achieves an acceptable success rate (80\% of the solved tasks), we stop training and proceed to the next stage. In the second stage, we re-train the policy in a new environment which is populated with static obstacles so the agent learns to avoid the collisions with them. In the last stage, we add an adversarial dynamic obstacle to the static environment so the agent learns to circumnavigate it or wait in place if needed to let the obstacle go away. The latter is the essential skill for planning with dynamic obstacles.

\subsection{Global planners}

\begin{algorithm}[t]
\caption{POLAMP with RRT planner}
\label{alg:RRT-POLAMP}
\begin{algorithmic}[1]
    \Require $s_{start}$, $s_{goal}$, $Obs(t)$, $N_{max}$, RL-PI, $D$, $N_{nbs}$, $R_{ex}$
    \Ensure $\mathcal{P}$: Motion Plan
    %\Function{POLAMP-RRT}{$s_{start}$, $s_{goal}, Obs$, RL-PI}
    \State $s_{start}.t \gets 0$
    \State $\mathcal{T} \gets$
    \Call{InitializeTree}{$s_{start}$}
    \While{$N_{\max}$ was not reached}
        \State $s_{rand} \gets$ \Call{RandomSample}{}
        \State $neighbors \gets$ \Call{Nearest}{$\mathcal{T}, s_{rand}$, $N_{nbs}$}
        \For{$s_i \in neighbors $}
            \State $s_j \gets$ \Call{Extend}{$s_i, s_{rand}, R_{ex}$}
            %\If{$\neg$ \Call{Collides}{$s_{to}, Obs(t)$}}
            \State $s_j \gets$ \Call{RL-Steer}{$s_i, s_j, s_{goal}, Obs(t)$, RL-PI, $D$}
            %\State \textcolor{blue}{$\triangleright$ $s_{t_n}.t$ is the time at which the agent reach $s_{t_n}$}
            \If{$s_j.tr$ is not empty}
                %\State \textcolor{blue}{$\triangleright$ the time $s_{t_n}.t$ is the current time of $s_{t_k}$ in A*}
                \State $\mathcal{T} \gets$ \Call{APPEND}{$s_j$}
                \If{$s_{goal}.tr$ is not empty}
                    \State $\mathcal{T} \gets$ \Call{APPEND}{$s_{goal}$}
                    \State \Return $\mathcal{P}$ = \Call{MotionPlan}{$\mathcal{T}$}
                \Else
                     \State \textbf{break}    
                \EndIf
            \EndIf
        \EndFor
%        \State $i \gets i + 1$
    \EndWhile
    \State \Return $\mathcal{P} = \emptyset$
    %\EndFunction
\end{algorithmic}
\end{algorithm}

%\begin{algorithm}[t]
%\caption{PolicySteer}
%\label{alg:Policy-Steer}
%\begin{algorithmic}[1]
%    \Require $s_{from}$: start state, $s_{to}$: goal state, $s_{global}$: planner goal state, $Obs(t)$, RL-PI
%    \Ensure $s_{to}, s_{goal}$: the states with their built trajectories
%    \State $tr \gets$ \Call{RL-PI}{$s_{from}, s_{to}, Obs(s_{from}.t)$}
%    \If{trajectory $tr$ is empty}
%        \State \Return $s_{to}, s_{goal}$
%    \EndIf
%    \State $s_{to}.t \gets s_{from}.t + tr.t$, $s_{to}.tr \gets tr$
%    \If{\Call{EuclDist}{$s_{to}, s_{goal}$} $< D$}
%        \State $tr_{goal} \gets$ \Call{RL-PI}{$s_{to}, s_{goal}, Obs(s_{to}.t)$}
%        \If{trajectory $tr_{goal}$ is not empty}
%            \State $s_{goal}.t \gets s_{to}.t + tr.t$, $s_{goal}.tr \gets tr_{goal}$
            %\State \Return true, true
%        \EndIf
%    \EndIf
%    \State \Return $s_{to}, s_{goal}$
    %\State \Return $\mathcal{P} = \emptyset$
    %\EndFunction
%\end{algorithmic}
%\end{algorithm}

Although our learnable local planner can generate a trajectory between two nearby states it is not well-suited for constructing a long-term plans. Thus we suggest using a global planner as well that can consistently explore different regions of the workspace relying on the global observation and find the ways to reach the remotely located goals. In this work, we utilize the classical algorithms RRT and A* as the global planners. For the detailed explanation of these algorithms we refer the reader to the original papers, and now proceed with an overview.

The pseudocodes of both algorithms are given in Alg.~\ref{alg:RRT-POLAMP} and Alg.~\ref{alg:ASTAR-POLAMP} respectively. The main difference between the sampling-based (i.e. RRT) and the lattice-based (i.e. A*) algorithms is how to choose the state to extend and how to extend the given state. On the one hand RRT uses \textit{RandomSample} in the state space to grow the search tree randomly from the \emph{Nearest} state in the tree using \emph{Extend} to limit the maximum distance of the states that should be connected. On the other hand, A* does not choose a random sample, but rather uses a deterministic priority queue of states, OPEN, to choose which state to expand (extend). The OPEN queue is sorted in order of increasing $f$-values, where $f(s) = g(s) + \epsilon \cdot h(s)$ consists of two terms $g(s)$ and $h(s)$. $g(s)$ is the cost of the shortest path from the start state to the current one, and $h(s)$ is the heuristic estimate of the cost from $s$ to goal. Upon choosing a most promising state A* the next states (\emph{Successors}) using a fixed set of motion primitives through which the robot reaches the next states.

The major difference between these classical algorithms and POLAMP is that POLAMP explicitly reasons about time moments to take the dynamic obstacles into account while planning. Local planning is implemented with the \textit{RL-STEER} function. This function solves a local planning problem, defined by the two states $s_i$ and $s_j$. If the distance between $s_i$ and $s_{goal}$ is less than $D$ then the goal is attempted to be reached from $s_i$. To reach the target state the policy \textbf{RL-PI} is used which has an access to the information on how the dynamic obstacles move, i.e. $Obs(t)$. If \textbf{RL-PI} managed to connect the states, it returns the generated trajectory $s_j.tr$ and the time by which the target state is reached, i.e. $s_j.t$. Thus all the states in the search tree bear the information on their reaching time which is used while planning.

%This part is shown in Alg.~\ref{alg:Policy-Steer} and uses our learned policy \emph{RL-PI}.

In this work, we use a modified version of RRT, when at each iteration \emph{Nearest} gets several $N_{nbs}$ with the maximum radius of extend $R_{ext}$ and tries to generate trajectories to them until one of them is build.
Unlike the original algorithm A*, where the search ends when the goal state is expanded, in this work, the search ends as soon as the trajectory to the final state is found. To generate the successors we use the technique of online motion primitives from~\cite{lin2021search}, i.e. we apply discrete controls $\xi = (a, \gamma)$ for a period of $H$ to determine the robot's desired configurations. Then we use our learned policy to construct collision-free trajectories to these configurations.

% that are a combination of discrete steering angles and linear velocities, and the Eq.~\ref{eq:diffEquationsRobot} during a period of time $H$ to generate the successors. Because we use our learned policy to connect two states we only generate the last state of these primitives, but not all the trajectory.
%Also, unlike the original paper, where is need to use obligatorily different discrete accelerations to generate the wait actions to avoid dynamic obstacles, we use only linear velocity because our policy is able to decelerate by himself. 

%The common parameters that both planners require are the following: start state $s_{start}$, the goal state $s_{goal}$, the set of obstacles over times $Obs(t)$, the number of maximum iterations $N_{max}$, the learned policy \emph{RL-PI} and the distance from which the policy can try to reach the goal state $D$.

\begin{algorithm}[ht!]
\caption{POLAMP with A* planner}
\label{alg:ASTAR-POLAMP}
\begin{algorithmic}[1]
    \Require $s_{start}$, $s_{goal}$, $Obs(t)$, $T$, $\xi$, $D$, RL-PI, $N_{max}$
    \Ensure $\mathcal{P}$: Motion Plan
    \State CLOSED $\gets \emptyset$, OPEN $\gets \emptyset$
    \State $s_{start}.t \gets 0$, $ g(s_{start}) \gets 0$, $f(s_{start}) \gets h(s_{start})$ 
    \State OPEN $\gets$ \Call{Insert}{$s_{start}$}
    \While{OPEN is not empty or $N_{max}$ was not reached}
        \State $s_i \gets $ OPEN.POP(), CLOSED $\gets$ \Call{Insert}{$s_i$}
        \State SUCCESSORS $\gets$ \Call{GetNextStates}{$\xi, T$}
        \For{$s_j \in$ SUCCESSORS}
            %\State \textcolor{blue}{$\triangleright$ $g(s_{t_k})$ is the current time of $s_{t_k}$ in A*}
            \State $s_j \gets$ \Call{RL-Steer}{$s_i, s_j, s_{goal}, Obs(t)$, RL-PI, $D$}
            \If{$s_j.tr$ is empty}
                \State \textbf{continue}
            \EndIf
            \If{$s_{goal}.tr$ is not empty}
                \State CLOSED $\gets s_{t_{n}}$, CLOSED $\gets s_{goal}$ 
                \State \Return $\mathcal{P}$ = \Call{MotionPlan}{CLOSED}
            \EndIf
            \State $c(s_i, s_j) \gets $\Call{COST}{$s_{t_n}.tr$}
%            \If {$s_{t_{n}}$ was not visited before}
%                \State $g(s_{t_{n}})) \gets \infty$, $f(s_{t_{n}})) \gets \infty$ 
%            \EndIf
            \If{$g(s_j)$ is better than any previous one}
                \State OPEN $\gets$ \Call{Insert}{$s_j$}
            \EndIf
            %\If{$g(s_{t_{n}}) > g(s{t_{k}}) + c(s{t_{k}}, s_{t_{n}})$}
            %    \State $g(s_{t_{n}}) = g(s{t_{k}}) + c(s{t_{k}}, s_{t_{n}})$
            %    \State $f(s_{t_{n}}) = g(s_{t_{n}}) + \epsilon \cdot h(s_{t_{n}})$
            %    \State OPEN $\gets$ \Call{Insert}{$s_{t_{n}}$}
            %\EndIf
        \EndFor
    \EndWhile
    \State \Return $\mathcal{P} = \emptyset$
    %\EndFunction
\end{algorithmic}
\end{algorithm}

%\begin{figure}[t]
%    \centering
%    \includegraphics[width=0.22\textwidth]{Images/vs-success-rate.png}
        %\label{fig:Map1}
        %\caption{Map1}
%        \includegraphics[width=0.22\textwidth]{Images/vs-time-to-reach.png}
        %\label{fig:Map2}
        %\caption{Map2}
%    \caption{Comparison of the local planners(LLP vs ExpStab)}
%    \label{fig:comparisonLocalPlanners}
%\end{figure}

\section{Experimental Evaluation}

We evaluated POLAMP (and compared it with the competitors) in two types of environments: with static obstacles and with both static and dynamic obstacles. 

\subsection{Policy learning}

To train the policy we created a dataset of different tasks (start and goal states) in three types of environments: empty, static, dynamic.  Every of these environments had a size $40m \times 40m$. Each task was generated randomly in a way that the distance between the start and goal locations was in the interval of $[15, 30]$m, moreover the difference in orientations did not exceeded $\frac{\pi}{4}$. The task was considered solved if the agent reached the goal state with the Euclidean error $\epsilon_{\rho} \leq 0.3$ m and the orientation error $\epsilon_{\theta} \leq \pi/18$ rad with no collisions. %The dataset is split into the training and validation subsets. The training tasks are used to train the policy network, while the validation tasks are used to calculate the performance of this policy after the training.

To generate tasks in static environments we sampled $12$ fragments of size $40m \times 40m$ from the map depicted on Fig.~\Ref{fig:environments} on the left (Map1), which has the size of $100m \times 60m$. 
%The static environments consist of parts generated randomly from the Map1 -- see Fig.~\ref{fig:environments}. In total we generated 12 static environments. For every of these we use the same task generation, but in this case we guarantee that the task can be resolved. To this end, we run RRT algorithm using ExpStab~ \cite{Astolfi1999ESPOSQ} as steering function to find out if the task is correctly or not. Additionally, we add a constraint to the length $l_{max} = 60$ m of the resulted trajectory to avoid the tasks with big length.
For training in dynamic environments we populated the static environments with one adversarial dynamic obstacle. I.e. the start state of the dynamic obstacles and its trajectory were generated semi-randomly in such way that with a very high chance it will intersect the path of the agent and will force the latter to detour/wait. An illustration is given in Fig.~\ref{fig:learningEnvironment}.

Similarly to the train dataset we created a separate set of validation tasks. We used them to measure the progress of training, i.e. once in a while we evaluated the performance of the currently trained policy on the validation tasks. If the success rate (the fraction of the solved tasks) was lower than 80\% we continue learning, in the opposite case -- we stopped learning.

\begin{figure}[!t]
    % \centering
    % \includegraphics{}
    \includegraphics[width=0.5\textwidth]{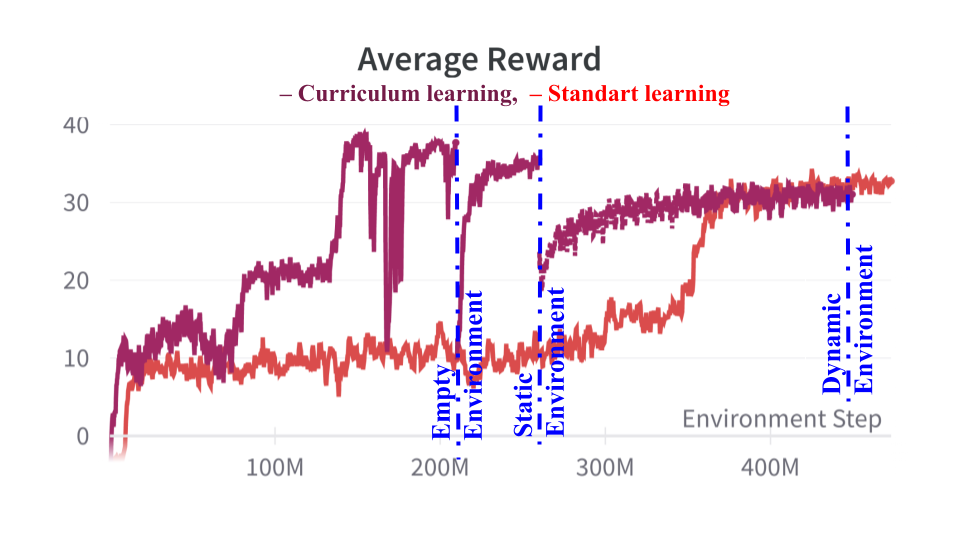}
    \caption{
    A comparison of learning curves between curriculum and standart learning for our policy. The dash lines represent the intermediate trained policy in the respecting environment.}
    \label{fig:curriculum-training}
\end{figure}

\textbf{The effect of curriculum learning.} To qualitatively assess the effect of the proposed curriculum learning we trained two policies: the first (baseline) was trained immediately in the dynamic environment, $\pi^{stand}$, while the second one, $\pi^{curr}$, was trained with the proposed three-stage curriculum. The corresponding learning curves are shown in Fig~\ref{fig:curriculum-training}. Evidently the curriculum policy $\pi^{curr}$ starts to converge from approx. 300M time step with almost 30 of reward and in this time the standard policy $\pi^{stand}$ only achieves the reward of 13 (and starts converging later). Thus, we confirm that the suggested curriculum leads to a faster convergence, which is especially useful when the resources, e.g. training time, are limited.

\begin{table}[ht]
\begin{center}
\resizebox{0.6\linewidth}{!}{
\begin{tabular}{p{0.10\linewidth}|p{0.10\linewidth}p{0.15\linewidth}p{0.10\linewidth}}
Agent & Dynamic & Orientation & SR \%\\
\hline \hline
$\pi^{st}_{w/o-\theta}$ & no & no & 99\\
$\pi^{st}_{w-\theta}$ & no & yes & 32\\
$\pi^{dyn}_{w/o-\theta}$ & yes & no & 28\\
$\pi^{dyn}_{w-\theta}$ & yes & yes & 22\\
\hline 
\hline
\end{tabular}}
\caption{The results of the trained DDPG agent in different setups.}
\label{tab:DDPG-agents}
\end{center}
\end{table}

\textbf{Training the learnable baseline.} The learnable baseline which we primarily aimed to compare with was RL-RRT~\cite{Chiang2019RL-RRT}. Similarly to POLAMP it is a combination of the global planner, RRT, with the learnable local planner, based on the DDPG policy. To provide a fair comparison we trained this policy on our dataset from scratch. However, even after a prolonged training its success rate on the validation tasks was not exceeding 22\%.

To understand the reasons of such performance we conducted additional training for the three variants of this policy in more simple setups. The characteristics of those setups and the resultant success rates are shown in Table~\ref{tab:DDPG-agents}. Notably, the policy that ignored the orientation constraints and dynamic obstacles (the same setting from RL-RRT), $\pi^{stat}_{w/o-\theta}$, showed a very good performance -- almost 100\% success rate. This goes in line with the original paper on RL-RRT as the authors considered this setting. However, when the setup becomes more complex, the performance of the policy drops significantly. For example, the policy which ignores the dynamic obstacle, $\pi^{stat}_{w/-\theta}$, showed only 32\% SR, and the one that ignores the goal orientation, $\pi^{dyn}_{w/o-\theta}$, -- 28\%. Thus, we conduct that this type of policy has an acceptable performance only in basic setups.
%Because in this article we propose a method that should work in dynamic environments with car-like constraints in the planning setup we used the policy $\pi^{dyn}_{w/-\theta}$. 

The poor performance of the RL-RRT in the case of more complex environmental conditions and with a large number of dynamic obstacles is primarily due to the instability of the learning process of the DDPG algorithm in a stochastic environment. DDPG belongs to the class of the off-policy methods, saves experience from different episodes in the replay buffer (including those that led to collisions), and generates a deterministic policy relative to the value function. In POLAMP, we use the on-policy PPO method, when only the latest relevant trajectories are considered to improve the policy, which at the later stages of training are unlikely to contain collision situations. In a number of works~\cite{SAC, MPO, Muesli}, on-policy algorithms showed a significant advantage over off-policy in a stochastic environment, due to the ability to generate a stochastic policy. The advantage of PPO over DDPG in our task is undeniable when using curriculum learning when a replay buffer prevents the DDPG from adapting to the new conditions of the next stage of training.

\begin{figure*}[t]
    \centering
    \includegraphics[width=0.31\linewidth]{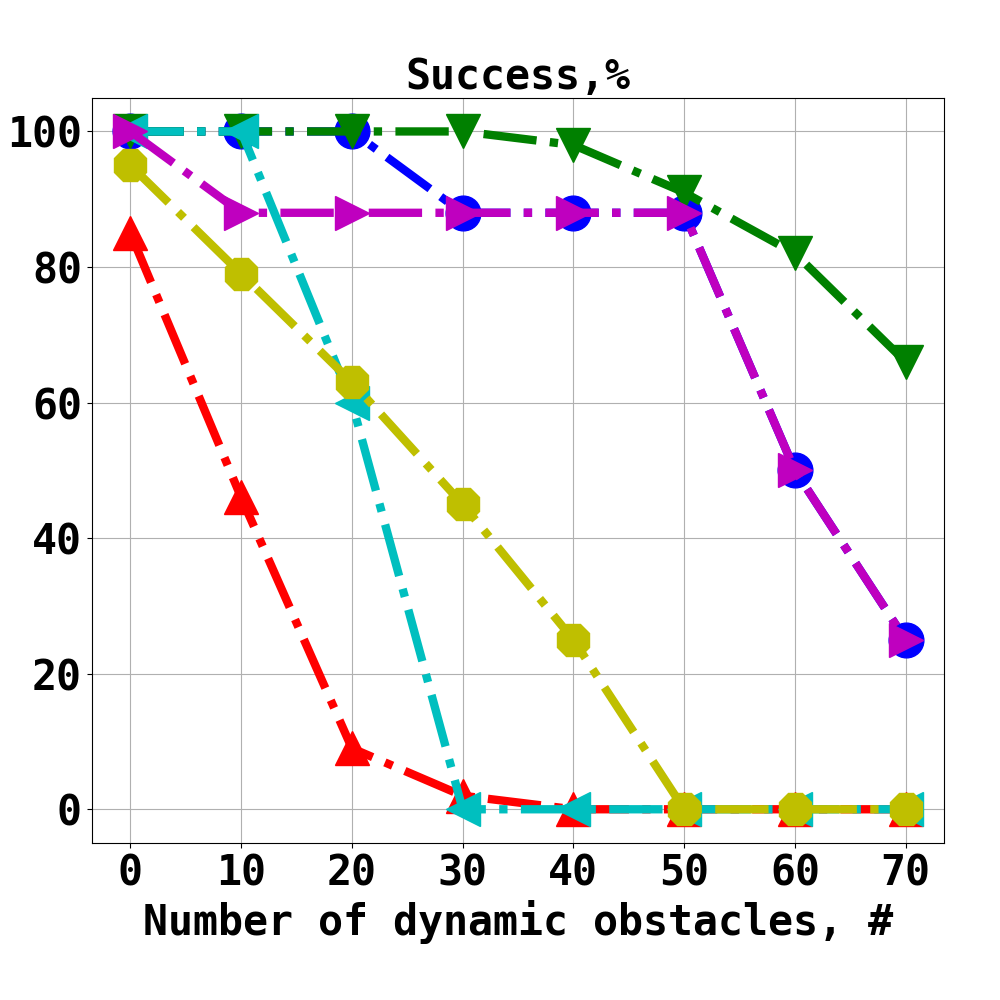}
    \includegraphics[width=0.31\linewidth]{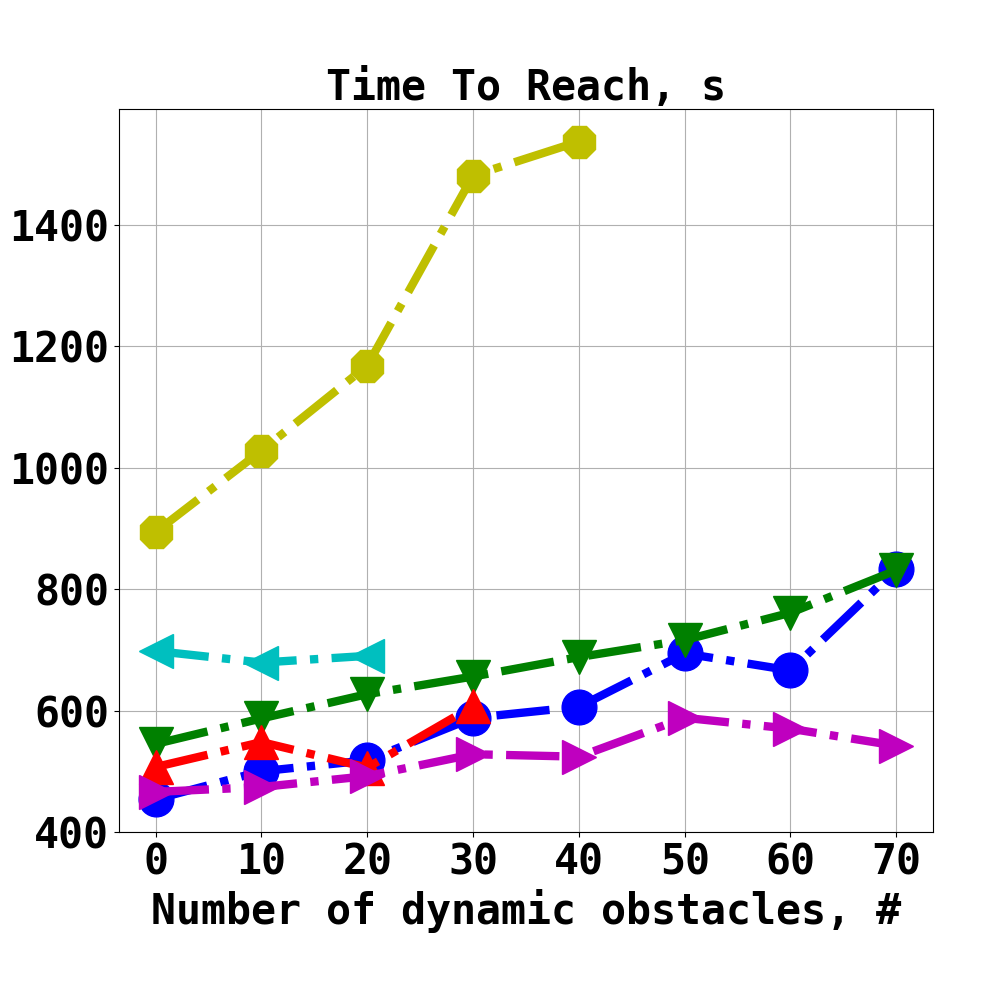}
    \includegraphics[width=0.31\linewidth]{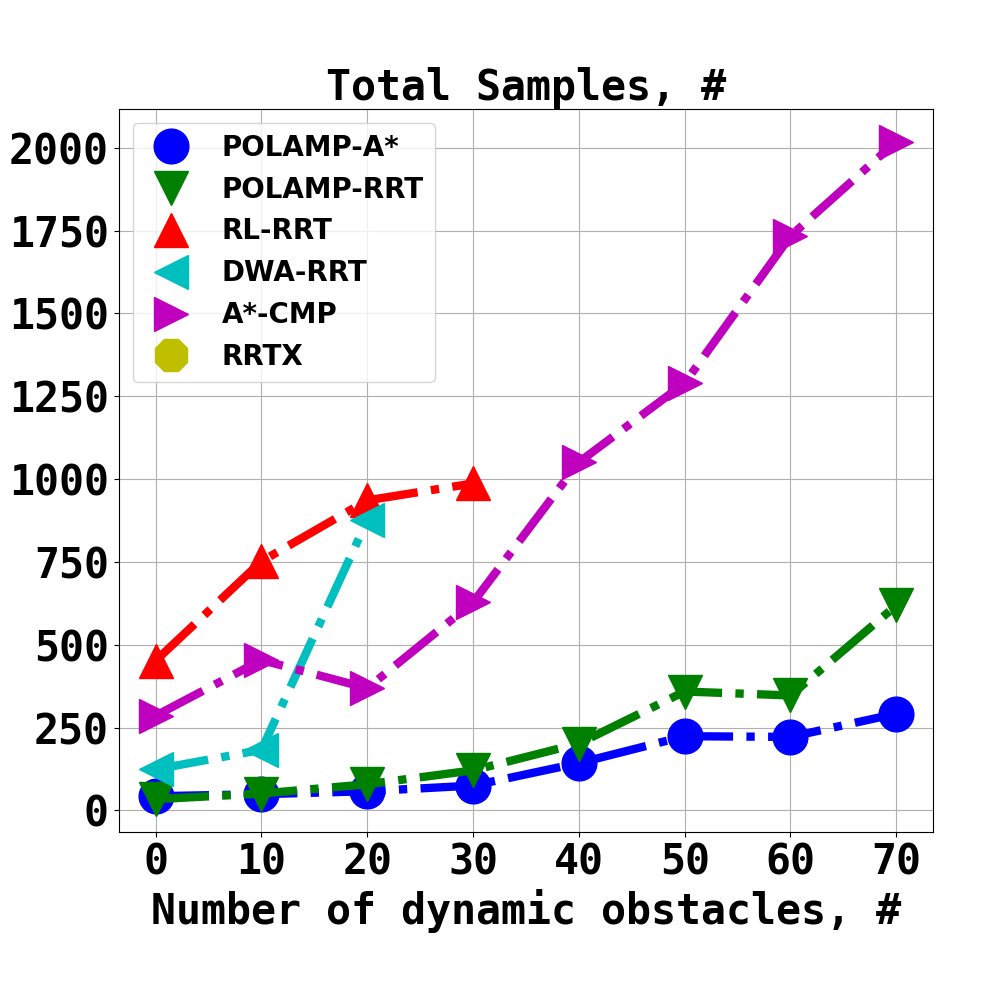}
    \caption{Planning results for the maps with dynamic obstacles (success rate, time to reach and number of samples). The legend for all algorithms is shown in the figure on the right.}
    \label{fig:MetricsOnDynamicMaps}
\end{figure*}

%Here $\rho$ is the Euclidean distance between the states $\sigma[i]$, $\sigma[j]$, denoted as $B$ and $G$ on Figure~\ref{fig:CarLike}. $\alpha$ is the angle of the goal vector with respect to the frame $B$, $\beta$ is the angle of the goal vector with respect to the $G$. $\omega$ is the rotational velocity. The latter can be transformed into the steering angle as follows: $\gamma = \arctan \dfrac{\omega \cdot L}{v}$.

%We compared our learning local planner and ExpStab individually and within the algorithm RRT. 

%\subsection{Comparison of the local planners}
\begin{table}[t]
\begin{center}
\resizebox{0.9\linewidth}{!}{
\begin{tabular}{p{0.05\linewidth}|p{0.25\linewidth}p{0.08\linewidth}p{0.08\linewidth}p{0.14\linewidth}p{0.11\linewidth}}
 
 Map & Planner & SR,\% & TTR,\% & Samples,\% & Time,\%\\
 \hline \hline
  \multirow{2}{4em}{1} & POLAMP-RRT & 100 & 100 & 100 & 100\\
  & POLAMP-A* & 100 & 93 & 179 & 103\\
  & RRT-ES & 90 & 120 & 3851 & 104\\
  & RL-RRT & 40 & 96 & 2424 & 578\\
  & SST* & 85 & 140 & 4124 & 111\\
 \hline 
  \multirow{2}{4em}{2} & POLAMP-RRT & 100 & 100 & 100 & 100\\
  & POLAMP-A* & 100 & 78 & 121 & 85\\
  & RRT-ES & 62.5 & 143 & 1322 & 107\\
  & RL-RRT & 4.5 & 153 & 677 & 308\\
  & SST* & 82.5 & 123 & 1225 & 102\\
 \hline
  \multirow{2}{4em}{3} & POLAMP-RRT & 100 & 100 & 100 & 100\\
  & POLAMP-A* & 100 & 84 & 143 & 89\\
  & RRT-ES & 31 & 102 & 3426 & 98\\
  & RL-RRT & 8 & 126 & 1532 & 407\\
  & SST* & 58.8 & 141 & 3560 & 101\\
 \hline
 \hline
\end{tabular}
}
\caption{Results of the experiments on the static maps.}
\label{tab:resultsForStaticEnvironment}
\end{center}
\end{table}

\subsection{Evaluation In Static Environments}

We used three different maps, resembling the parking lots, for the evaluation -- see Fig.~\ref{fig:environments}. Each map had a size of  $100m \times 60m$ and was generated based on the dataset from~\cite{parkingDataSet}. Please note, that only several fragments of Map1 were observed by the policy during training, while Map2 and Map3 were not used while training at all. For each map, we generated 20 different planning instances, i.e. the start-goal location pairs. We generated them randomly and discarded the instances for which the straight-line distance between start and goal was less than $50$m (in order to avoid non-challenging tasks). Start/goal orientations were also chosen randomly as the multiplicative of $90^\circ$. Each test was repeated 30 times. A test was counted as failure if the robot was not able to reach the goal with following tolerance: $\epsilon_{\rho} \leq 0.5$ m and $\epsilon_{\theta} \leq \pi / 18$.

We compared POLAMP to the following algorithms: RRT that utilized a well-known non-learnable steering function based on the exponential stabilization~\cite{Astolfi1999ESPOSQ} (denoted RRT-ES), a kinodynamic motion planner SST*~\cite{li2016asymptotically}, RL-RRT~\cite{Chiang2019RL-RRT} -- a state-of-the-art planning method with a learnable local planner (details on learning this planner were provided above). 

For the RRT part of the algorithms, we set the radius of the Extend method $R_{ext} = 10$ m, the maximum distance which we can reach the goal from is $D = 30$ m, the number of nearest neighbors $N_{nbs} = 5$  and the maximum number of iterations of the RRT $N_{max} = 1500$ for the POLAMP-RRT and RL-RRT, and $N_{max} = 3000$ for the RRT-ES and SST*. For POLAMP-A* we used the same parameters as for RRT. Additionally, we used the 7 discrete steering angles ranged uniformly between $[\gamma_{min}, \gamma_{max}]$, the linear velocity $v=2$ and the time horizon $H = 3$ s to generate the lattice of the motion primitives. All these values were chosen following a preliminary evaluation aimed at identifying the suitable parameters' values.

The metrics we used were: success rate (SR) -- how often the planner produces a path that reaches the goal, time to reach the goal (TTR), total number of samples and the runtime of the algorithm.

The results are presented in Table~\ref{tab:resultsForStaticEnvironment}. Notably, POLAMP has a much higher success rate compared to the other algorithms reaching almost 100\% in every map. This shows that our learnable local planner, indeed, generalizes well to the unseen consitions. The observable trend is that POLAMP requires much fewer samples than RRT-ES to generate the motion plan. For example, for Map2 POLAMP requires 14x and 12x less samples compared to RRT-ES and SST* respectively. This is because POLAMP performs collision avoidance for local steering while RRT-ES and SST* do not. In comparison with RL-RRT, POLAMP also requires less samples.

Also, is can be noted that RL-RRT has a higher success rate for the Map1 than for the other maps, meaning that, unlike our policy, the policy of RL-RRT did not generalize well to the other two maps. We can suggest that the main reason for RL-RRT not being able to perform well on Map2 and Map3 is that the learnable component of that planner, i.e. DDPG, was not able to learn sufficiently well in our setup, i.e. provided only with the instances that were taken from the Map1. In other words, the DDPG policy  was not able to learn well in our dataset and was overffited to Map1. Meanwhile, PPO that used the same amount of data for training, was able to generalize to solving local pathfinding queries on (the unseen during training) Map2 and Map3. Thus, we infer that PPO is a more sample efficient policy that, generally, should be preferred over DDPG in similar setups.

\subsection{Evaluation In Dynamic Environments}

%Each obstacle was moving randomly in the environment. 
For this series of the experiments, we used Map2 and Map3, i.e. the maps that were not used for training. These maps were populated with the varying number of dynamic obstacles: from $0$ to $70$. Every dynamic obstacle is a rectangular shape car-like robot. Its trajectory is generated by sampling the random control input $(a, \omega)$ every 10th time step. We generated 5 different trajectories for every dynamic obstacle. Two different start-goal pairs were chosen for each map. Each test was repeated 20 times for the sampling-based planners.

As before we compared POLAMP to RL-RRT. We also compared to A*-CMP~\cite{lin2021search}. For this algorithm we used the same parameters as for POLAMP-A*%, but used two discrete accelerations $a_i = {-5, 5} m/s^2$ instead of linear velocity
. Another baseline was the combination of RRT with the seminal Dynamic Window Approach (DWA)~\cite{DWA} as a local planner (RRT-DWA). The latter is capable of avoiding moving obstacles and is widely used in robotics. For RRT-DWA we did not account for the final orientation as DWA is not tailored to obey orientation constraints. Also, we compared to RRTX~\cite{Otte2016RRTX} that used Dubins steering function~\cite{DUBINS}. This algorithm is essentially a plan-execute-re-plan type of algorithm that re-uses the search tree while the robot is moving towards the goal. For better performance of RRTX at each re-planning iteration we did not take into account the moving obstacles located more then 20 m away from robot. In the case of RRTX, the SR means how often the robot can reach the goal without collisions while executing the path. Additionally because RRTX needs much more samples during the re-planning we do not show this metric for RRTX.

The results are presented in Fig~\ref{fig:MetricsOnDynamicMaps}. The first clear trend is that POLAMP-RRT, POLAMP-A* and A*-CMP
in all cases maintain a high success rate ($>92\%$) until the number of dynamic obstacles goes beyond 50. However, POLAMP-A* and POLAMP-RRT require much fewer samples than A*-CMP to find the trajectory. This is because the A*-CMP requires two groups of primitives. One group of primitives allows accelerate and move at a constant speed while another group tries to decelerate to avoid collision with dynamic obstacles. However our algorithm only requires one group of primitives, because our policy is able to decelerate to avoid collision with dynamic obstacles when it is necessary.

We also note that there are trade-off between POLAMP-RRT, POLAMP-A* and A*-CMP. On the one hand, POLAMP-RRT is slightly better than the baseline A*-CMP and our POLAMP-A* in terms of success rate. Thanks to the randomness of RRT, POLAMP-RRT is able to explore more and can solve complicated tasks, unlike A* which performs a systematic non-explorative search. On the other hand, A*-CMP has the lowest duration in comparison with the rest algorithms. The latter is because in each iteration A*-CMP uses the minimum and maximum acceleration to generate the neighbors, i.e. the algorithm makes abrupt changes in speed. However, our local learnable steering tries to change the speed smoothly due to the presence of obstacles. Our algorithm is better than the other baselines RL-RRT, and RRT-DWA. Due the poor performance of the $\pi^{dyn}_{w-\theta}$ the RL-RRT algorithm did not show good results. RRT-DWA works well only when the number of obstacles is small. %But when it encounters more moving obstacles, it falls into local minima, from which it is no longer possible to avoid collision with obstacles.
%It is because DWA tries to avoid moving obstacles, but also tends to reach the next desired state, i.e. does not have the ability to wait.

POLAMP-RRT and POLAMP-A* are also better than RRTX. RRTX tries to replan the path online but sometimes when the current path is occluded by dynamic obstacles the robot is forced to stop and stay in its place until it finds another solution. In these situations, the robot can get into a deadlock from where it is impossible to get out without a collision because of moving obstacles. This problem is due to RRTX not taking into account the future trajectories of dynamic obstacles while planning. Besides, TTR of RRTX is almost double that of the other algorithms. This is because RRTX has abrupt path changes when the path is affected by the appearance of dynamic obstacles.

Overall, the conducted experiments show that our policy $\pi^{curr}$ generalizes well to both new environments and increasing number of dynamic obstacles (recall that it was trained only with one moving obstacle). A combination of that policy with a search-based or sampling-based global planner works well in challenging environments with dozens of simultaneously moving obstacles. Some experimental videos are provided in the Multimedia Materials.

\section{Conclusion}

In this paper, we considered a problem of kinodynamic planning for non-holonomic robot in the environments with dynamic obstacles. We enhanced the two classical planning methods, A* and RRT, with a learnable steering function that takes into account kinodynamic constraints and both static and moving obstacles. We designed a reward function and created a specific curriculum for learning the steering behaviors. The resultant algorithm, POLAMP, was evaluated empirically in both static and dynamic environments and was shown to outperform the state-of-the-art baselines (both learnable and non-learnable).

%\noindent Congratulations 
\bibliographystyle{IEEEtran}
\bibliography{IEEEexample}

\end{document}